\documentclass[final]{cvpr}
\usepackage[T1]{fontenc}
\usepackage{times}
\usepackage{epsfig}
\usepackage{graphicx}
\usepackage{amsmath}
\usepackage{amssymb}
\usepackage{float}
\usepackage{caption}
\usepackage{subcaption}
\usepackage{soul}
\usepackage{booktabs}
\usepackage{mathtools}
\usepackage{appendix}
\usepackage[ruled,linesnumbered]{algorithm2e}
\usepackage{esvect}
\SetKwInOut{Input}{Input}\SetKwInOut{Output}{Output}\SetKwInOut{Parameter}{Parameter}
\usepackage[pagebackref=true,breaklinks=true,colorlinks,bookmarks=false]{hyperref}

\def\cvprPaperID{3079} 
\def\confYear{CVPR 2021}

\begin{document}
 
\title{Noise-resistant Deep Metric Learning with Ranking-based Instance Selection}

\author{%
 \textbf{Chang Liu}\textsuperscript{1}, \quad \textbf{Han Yu}\textsuperscript{1}, \quad \textbf{Boyang Li}\textsuperscript{1*}, \quad \textbf{Zhiqi Shen}\textsuperscript{1*},\quad  \textbf{Zhanning Gao}\textsuperscript{2*},\\  \textbf{Peiran Ren}\textsuperscript{2},\quad  \textbf{Xuansong Xie}\textsuperscript{2}, \quad  \textbf{Lizhen Cui}\textsuperscript{3,4}, \quad  \textbf{Chunyan Miao}\textsuperscript{1*}\\
\textsuperscript{1}School of Computer Science and Engineering, Nanyang Technological University (NTU), Singapore\\
\textsuperscript{2}Alibaba Group, Hangzhou, China \quad \textsuperscript{3}School of Software, Shandong University (SDU), Jinan, China\\
\textsuperscript{4}Joint SDU-NTU Centre for Artificial Intelligence Research (C-FAIR), SDU, Jinan, China\\
*\{boyang.li, zqshen\}@ntu.edu.sg, zhanning.gzn@alibaba-inc.com, ascymiao@ntu.edu.sg}

\maketitle

\begin{abstract}
The existence of noisy labels in real-world data negatively impacts the performance of deep learning models. Although much research effort has been devoted to improving robustness to noisy labels in classification tasks, the problem of noisy labels in deep metric learning (DML) remains open. In this paper, we propose a noise-resistant training technique for DML, which we name Probabilistic Ranking-based Instance Selection with Memory (PRISM). PRISM identifies noisy data in a minibatch using average similarity against image features extracted by several previous versions of the neural network. These features are stored in and retrieved from a memory bank. To alleviate the high computational cost brought by the memory bank, we introduce an acceleration method that replaces individual data points with the class centers. In extensive comparisons with 12 existing approaches under both synthetic and real-world label noise, PRISM demonstrates superior performance of up to 6.06\% in Precision@1. 
\end{abstract}

\section{Introduction}
Commonly resulting from human annotation errors or imperfect automated data collection, noisy labels in training data degrade the predictive performance of models trained on them \cite{han2018co,wang2018iterative,jiang2020beyond}. Manual inspection and correction of labels are labour-intensive and hence scale poorly to large datasets. Therefore, training techniques that are robust to incorrect labels in training data play an important role in real-world applications of machine learning.  



To date, most works on noise-resistant neural networks \cite{han2018co,jiang2018mentornet,patrini2017making,yu2019does,wang2018iterative,wei2020combating,jiang2020beyond} focus on image classification. Little research effort has been devoted to noise-resistant deep metric learning (DML).
The goal of DML is to learn a distance metric that maps similar pairs of data points close together and dissimilar pairs far apart, based on a predefined notion for similarity. DML finds diverse applications such as image retrieval \cite{kaya2019deep,gordo2017end,revaud2019learning}, landmark identification \cite{weyand2020google}, and self-supervised learning \cite{miech2019endtoend}. 


Pair-based loss functions encourages DML networks to distinguish a similar pair of data points from one or more dissimilar pairs. Large batch sizes often lead to improved performance \cite{cakir2019deep,wang2020cross,brown2020smooth}, as larger batches are more likely to contain informative examples. Pushing the idea of large batches to an extreme, \cite{wang2020cross} collects all positive and negative data samples from a memory bank. However, in the presence of substantial noise, indiscriminate use of all samples could lower performance. Alternatively, \cite{movshovitz2017no} uses learnable class centers to replace individual data samples in order to reduce computational complexity. Nonetheless, the cluster centers can also be sensitive to outliers and label noise. 

We propose a noise-resistant deep metric learning algorithm, \emph{Probabilistic Ranking-based Instance Selection with Memory} (PRISM), which works with both the memory bank approach and the class-center approach. PRISM computes the probability that a label is clean based on the similarities between the data point and other data points using features extracted during the last several training iterations. This may be seen as modeling the posterior probability of the data label. For data points with high probability, we extract their features and insert them into the memory bank, which is used in subsequent model updates. 
In addition, we develop a smooth top-$R$ (sTRM) trick to adjust the threshold for noisy data identification as well as an acceleration technique that replaces individual data points with the class centers in the probability calculation. 

We perform extensive empirical evaluations on both synthetic and real datasets. Inspired by the the ``noise cluster'' phenomenon observed from real-world data, we introduce the Small Cluster noise model to mimic open-set noise in real data. Experimental results show that PRISM achieves superior performance compared to 12 existing DML and noise-resistant training techniques under symmetric noise, Small Cluster noise, and real noise. In addition, the acceleration trick speeds up the algorithm by a factor of 6.9 on SOP dataset. The code and data are available at \url{https://github.com/alibaba-edu/Ranking-based-Instance-Selection}.

\section{Related Work}
\noindent \textbf{Noise-resistant Training in Classification.} Training under noisy labels has been studied extensively for classification \cite{angluin1988learning,wang2018iterative,wei2020combating,jiang2020beyond,zheng2021mlc,algan2020meta}. A common approach is to gradually detect noisy labels and exclude them from the training set. F-correction \cite{patrini2017making} models the noise as a class transition matrix. MentorNet \cite{jiang2018mentornet} trains a teacher network that provides weight for each sample to the student network. Co-teaching \cite{han2018co} trains two networks concurrently. Samples identified as having small loss by one network is used to train the other network. Co-teaching+ \cite{yu2019does} trains on samples that have small losses and different predictions from the two networks. 

Differing from the conventional classification problem, metric learning learns an effective distance metric that works well for unseen classes and is evaluated using retrieval-based criteria. In the experiments, PRISM demonstrates superior performance to several noise-resistant classification baselines. 

\noindent \textbf{Models of Label Noise.} Though noisy labels are prevalent, it is often difficult to ascertain and control the degree of noise in natural datasets. Artificial noisy models are thus commonly used as evaluation metrics for noise-resistant algorithms. In the symmetric noise model \cite{van2015learning}, a proportion $R$ of data points belonging to one ground-truth class are uniformly distributed to all other classes. In pairwise noise (e.g., \cite{han2018co}), data points from each class are transferred to a designated target class. \cite{jiang2020beyond} curates a natural dataset with controlled levels of noise.

Open-set noise refers to the presence of data points that do not belong to any classes recognized by the dataset. 
Under open-set noise, it is futile to model noise as a class transition matrix, which adds to the difficulty.  In classification, \cite{wang2018iterative} simulates open-set noisy labels by adding data from other datasets. In this paper, we propose the Small Cluster noise model for open-set label noise in metric learning. 

\noindent \textbf{Deep Metric Learning.} 
We broadly categorize deep metric learning into 1) pair-based methods and 2) proxy-based methods. 
Pair-based methods \cite{schroff2015facenet,wang2019ranked,wang2019multi,sun2020circle} calculate loss based on the contrast between positive pairs and negative pairs, which is often calculated using contrastive loss \cite{chopra2005learning}, triplet losses \cite{hermans2017defense}, or softmax loss \cite{goldberger2004neighbourhood}. In this process, identifying informative positive and negative pairs becomes an important consideration \cite{Song2016:lifted-embedding,harwood2017smart,Suh_2019_CVPR,cakir2019deep,wang2019multi,sun2020circle,wu2017sampling}. Proxy-based methods \cite{movshovitz2017no,qian2019softtriple,cakir2019deep,kim2020proxy,teh2020proxynca++,zhu2020fewer,elezi2020group,zhaiclassification,boudiaf2020unifying} represent each class as one or more proxy vectors, and use the similarities between the input data and the proxies to calculate the loss. Proxies are learned from data during model training, which could deviate from the class center under heavy noise and cause performance degradation. 


\noindent \textbf{Noisy Labels in Metric Learning:} To our knowledge, the only method which explicitly handles noise in neural metric learning is \cite{wang2017robust}. The technique estimates the posterior label distribution using a shallow Bayesian neural network with only one layer. Due to its computational complexity, the approach may not scale well to deeper network architectures. 

A few works attempt to handle outliers in normal training data in DML, but do not explicitly deal with substantial label noise. Wang \etal\cite{wang2019deep} uses the pair-based loss and trains a proxy for each class simultaneously to adjust the weights of the outliers, but substantial label noise may cause the learned proxies to be inaccurate.
Ozaki \etal\cite{ozaki2019large} handles noisy data in DML by first performing label cleaning using a model trained on a clean dataset, which may not be available in real-world applications. In this paper, we do not rely on the existence of a clean dataset.


\section{The PRISM Approach}

The detection of noisy data labels usually attempts to catch data points that stand out from others in the same class. However, distinguishing such data samples in DML requires a good similarity metric. The learning of a good similarity metric in turn depends on the availability of clean training data, thereby creating a chicken-and-egg problem. 

To cope with this challenge, PRISM adopts an online data filtering approach. At every training iteration, we use the features extracted during the past several iterations to filter out a portion of training data. The rest are considered clean, added to the memory bank, and used in updating the metric.

\subsection{Identifying Noisy Labels}
Let the training set be $\mathcal{X}=\{(x_0,y_0), (x_1,y_1), ...,$ $(x_N,y_N)\}$, where $x_i$ is an image and $y_i$ is the corresponding label. $N$ is the size of training set. The aim of DML is to learn a convolutional neural network, $f(\cdot)$, which extracts a feature vector $f(x_i)$ for image $x_i$, such that the cosine similarity $S(f(x_i), f(x_j))$ between $f(x_i)$ and $f(x_j)$ is high if $y_i=y_j$ and low if $y_i \neq y_j$: 
\begin{equation}
S(f(x_i), f(x_j))=\frac{f(x_i)^Tf(x_j)}{\|f(x_i)\| \|f(x_j)\|}.
\end{equation}
In addition, we denote the current mini-batch as $\mathcal{B}=\{(x_0,y_0),(x_1,y_1),...,(x_B,y_B)\}$ where $B$ is the batch size. A pair of feature $(f(x_i), f(x_j))$ is called positive pair if $y_i=y_j$, negative pair if $y_i \neq y_j$.

To identify noisy labels, we maintain a first-in first-out memory bank $\mathcal{M}$, $\mathcal{M}=\{(v_0,y_0),(v_1,y_1),...,(v_M,y_M)\}$, to store historic features of data samples. $M$ is the size of memory bank. In every step of stochastic gradient descent, we separate the clean data from the noisy data, and append the current feature $v_i$ of clean data $x_i$ to the memory bank. If the maximum bank capacity is exceeded, the oldest features are dequeued from the memory bank, so that we always keep track of the more recent features.


We compare the features of $x_i$ with the content of the memory bank to determine if its label $y_i$ is noisy.
If $y_i$ is a clean label, then the similarity between $x_i$ and other samples with the same class label in the memory should be large compared to its similarity with samples from other classes. Based on this intuition, we define the probability of $(x_i, y_i)$ being a clean data point, $P_{\text{clean}}(i)$, as follows
\begin{equation}\label{equ:pi}
P_{\text{clean}}(i) = \frac{\exp\left(T(x_i, y_i)\right)}{\sum_{k \in C}\exp \left(T(x_i, k\right))}
\end{equation}
\begin{equation}
T(x_i, k) = \frac{1}{M_k}\sum\limits_{(v_j,y_j) \in \mathcal{M}, y_j=k}S(f(x_i),v_j)
\end{equation}
$M_k$ is the number of samples in class $k$ in the memory bank. $T(x_i, k)$ is the average similarity between $x_i$ and all the stored features $v_j$ in class $k$. $T(x_i, k)$ may be seen as $\log P(X=x_i|Y=k)$ up to a constant. Equation~\ref{equ:pi} can be understood as the probability $P(Y=k|X=x_i)$, assuming a uniform prior for $P(Y=k)$ and identical normalization constants for every class. Although similar math forms can be found in applications such as metric learning \cite{goldberger2004neighbourhood,qian2019softtriple}, data visualization \cite{vandermaaten2008:tsne}, and uncertainty estimation \cite{mandelbaum2017distance,xing2020distance}, we note that its use in noise-resistant DML is novel. 

When $P_{\text{clean}}(i)$ falls below a threshold $m$, we treat $(x_i, y_i)$ as a noisy data sample. 
We propose two methods to determine the value of threshold $m$: the top-$R$ method (TRM) and the smooth top-$R$ method (sTRM). Under TRM, we define a filtering rate (\emph{i.e.}, estimated noise rate) $R$. In each minibatch, we treat $(x_i, y_i)$ as noisy if $P_{\text{clean}}(i)$ falls in the smallest $R\%$ of all samples in the current minibatch $\mathcal{B}$. 

In contrast, sTRM keeps track of the average of the $R^{\text{th}}$ percentile of $P_{\text{clean}}(i)$ values over the last $\tau$ batches. Formally, let $Q_j$ be the $R^{\text{th}}$ percentile $P_{\text{clean}}(i)$ value in $j$-th mini-batch, the threshold $m$ is defined as:
\begin{equation}\label{equ:sTRM}
m=\frac{1}{\tau}\sum_{j=t-\tau}^{t}Q_j
\end{equation}
Compared to TRM, the sliding window approach of sTRM reduces the influence of a single mini-batch and creates a smooth and more accurate estimate of the $R^{\text{th}}$ percentile. 

To create balanced minibatches, we first sample $P$ unique classes and sample $K$ images for each selected class, yielding $PK$ images in every minibatch.

\subsection{Accelerating PRISM}
The above method requires computing the similarities between all pairs of data samples, which has high time complexity. We propose a simple technique for improving the efficiency. For the $k^{\text{th}}$ cluster, we replace its $M_k$ data samples with the mean feature vector $w_k$ of the class,

\begin{align}
\begin{split}\label{equ:wk}
    \sum_{\substack{(v_j,y_j) \in \mathcal{M} \\ y_j=k}}\frac{S(f(x_i),v_j)}{M_k}&=\left(\frac{1}{M_k}\sum_{\substack{(v_j,y_j) \in \mathcal{M} \\ y_j=k}}\frac{v_j}{\|v_j\|}\right)\frac{f(x_i)}{\|f(x_i)\|} \\
&=w_k\frac{f(x_i)}{\|f(x_i)\|},
\end{split}\\
\begin{split}\label{equ:wk_used_in_alg1}
    w_k &= \frac{1}{M_k} \sum_{(v_j,y_j) \in \mathcal{M},  y_j=k} \frac{v_j}{\|v_j\|}.
\end{split}
\end{align}




Plugging Eq.~\eqref{equ:wk} into Eq.~\eqref{equ:pi}, $P_{\text{clean}}(i)$ can be expressed as:
\begin{equation}\label{equ:pi2}
P_{\text{clean}}(i)=\exp\left(w_{y_i}\frac{f(x_i)}{\|f(x_i)\|}\right) / {\sum_{k \in C}\exp\left(w_k\frac{f(x_i)}{\|f(x_i)\|}\right)}.
\end{equation}
The time complexity of Eq.~\eqref{equ:pi} is $O(PKN)$ for a minibatch of size $PK$ and a training set of $N$ samples. Adopting Eq.~\eqref{equ:pi2} reduces the time complexity to $O(PK|C|)$, where $|C|$ is the total number of classes which is much smaller than $N$. This yields an acceleration factor of $\frac{N}{|C|}$.

\subsection{The PRISM Algorithm}

We now describe the PRISM algorithm, shown as Algorithm~\ref{alg:nsm}. In each iteration of training, PRISM first calculates $P_{\text{clean}}(i)$ for each $(x_i,y_i)$ tuple. In the first iteration, when the class center vector $w_{y_i}$ has not been updated and is the zero vector, all data samples in class $y_i$ are considered clean (Line 3). At this moment, the memory bank does not contain any data points of class $k$, so we cannot compute the mean vector $w_k$. After the first iteration, we will update the center vector to a non-zero value and perform sample selection based on $P_{\text{clean}}(i)$. 

After that, we compute the threshold $m$ and add 
data points with $P_{\text{clean}}(i) > m$ to $\mathcal{B}_{clean}$ (Line 12).
To reduce computational cost, we update only the center vectors of classes in $\mathcal{B}_{clean}$ (Line 17). Finally, the loss is calculated to update parameters of model $f(\cdot)$ (Line 19). PRISM computes the loss with clean labels and perform stochastic gradient descent. The loss functions are described in the following section. 


\subsection{Loss Functions}
The traditional pair-based contrastive loss function \cite{chopra2005learning} computes similarities between all pairs of data samples within the mini-batch $\mathcal{B}$. The loss function encourages $f(\cdot)$ to assign small distances between samples in the same class and large distances between samples from different classes. 
More formally, the loss for mini-batch $\mathcal{B}$ is
\begin{equation}
\begin{split}
    L_{\text{batch}}(\mathcal{B}) & =\sum_{\mathclap{\substack{(x_i, y_i) \in B, (x_j, y_j) \in B\\ y_i \neq y_j}}} \max(S(f(x_i), f(x_j))    -\lambda,0) \\
    &-\sum_{\mathclap{\substack{(x_i, y_i) \in B, (x_j, y_j) \in B\\ y_i = y_j}}} S(f(x_i), f(x_j))
\end{split}
\end{equation}
where $\lambda\in[0,1]$ is a hyperparameter for the margin.
With a memory bank $\mathcal{M}$ that stores the features of data samples in previous minibatches in a first-in-first-out manner \cite{wang2020cross}, we can employ many more positive and negative pairs in the loss, which may reduce the variance in the gradient estimates. The memory bank loss can be written as:
\begin{equation}
\begin{split}
    L_{\text{bank}}(\mathcal{M}, \mathcal{B}) & =\sum_{\mathclap{\substack{(x_i,y_i) \in \mathcal{B},\,  (v_j,y_j) \in \mathcal{M}\\ y_i \neq y_j}}} \max(S(f(x_i), v_j) -\lambda,0)\\ 
    & -\sum_{\mathclap{\substack{(x_i,y_i)  \in \mathcal{B},\, (v_j,y_j) \in \mathcal{M} \\ y_i = y_j}}} S(f(x_i),v_j).
    \end{split}
\end{equation}

\begin{algorithm}[t!]
	\SetAlgoLined
	\Input{$\mathcal{B}=\{(x_0,y_0),(x_1,y_1),...,(x_B,y_B)\}$: a given minibatch of data with size $B$;\\$f(\cdot)$: a given deep metric model;}
	\Parameter{$\{w_k|k\in C\}$: the set of mean feature vectors of all classes,
	all initialized to zero before training commences;}
	\BlankLine
	\For{\textup{\textbf{each}} $(x_i,y_i) \in \mathcal{B}$}{
	    \eIf{$w_{y_i}=\vv{0}$}
	    {
	        $P_{\text{clean}}(i)=1$;
	    }
	    {
	        Calculate $P_{\text{clean}}(i)$ according to Eq.~\eqref{equ:pi2};
	    }
	}
	Calculate the threshold $m$ using TRM or sTRM;\\
	Initialize $\mathcal{B}_{clean}$ as an empty set;\\
	\For{\textup{\textbf{each}} $(x_i,y_i) \in \mathcal{B}$}{
	    \If{$P_{\text{clean}}(i)>m$}
	    {
	        Add $(f(x_i),y_i)$ to $\mathcal{B}_{clean}$;\\
	    }
	}
	Enqueue $\mathcal{B}_{clean}$ into the Memory Bank\\
	\For{\textup{\textbf{each}} $(v_i,y_i) \in \mathcal{B}_{clean}$}{
	    Update $w_{y_i}$ according to Eq.~\eqref{equ:wk_used_in_alg1};
	}
	Calculate loss $L(\mathcal{B}_{clean})$ and update the parameters of $f(\cdot)$\\
	
	\BlankLine
	\caption{A training iteration of PRISM}\label{alg:nsm}
\end{algorithm}

The total loss is the sum of the batch loss $L_{\text{batch}}(\mathcal{B})$ and the memory bank loss $L_{\text{bank}}(\mathcal{M}, \mathcal{B})$, referred to as the \emph{memory-based contrastive loss} \cite{wang2020cross}. As we adopt the memory bank setup to identify data with noisy labels, PRISM works well under the memory-based contrastive loss. 

Another loss we employ with PRISM is the Soft Triple loss \cite{qian2019softtriple}, a type of proxy-based loss function. This loss maintains $H$ learnable proxies per class. A proxy is a vector that has the same size with the feature of an image. The similarity between an image to a given class of images is represented as a weighted similarity to each proxy in the class. The loss is computed as the similarities between the minibatch data and all classes:
\begin{equation}\label{eq:softtriple}
    L_{\text{SoftTriple}}=-\log\frac{\exp(\lambda(S'_{i,y_i}-\delta))}{\exp(\lambda(S'_{i,y_i}-\delta))+\exp(\lambda S'_{i,j})},
\end{equation}
\begin{equation}
    S'_{i,j}=\frac{\sum_{h=1}^H\exp\left(\gamma f(x_i)^\top p_j^h\right)f(x_i)^\top p_j^h}{\sum_{h=1}^H \exp\left(\gamma f(x_i)^\top p_j^h\right)}.
\end{equation}
$\lambda$ and $\gamma$ are predefined scaling factors. $\delta$ is a predefined margin. $p_j^h$ is the $h$-th proxy for class $j$, which is a learnable vector updated during model training.

\section{Experimental Evaluation}

\subsection{Datasets}
\begin{figure}[t]
      \centering
      \includegraphics[width=1\columnwidth]{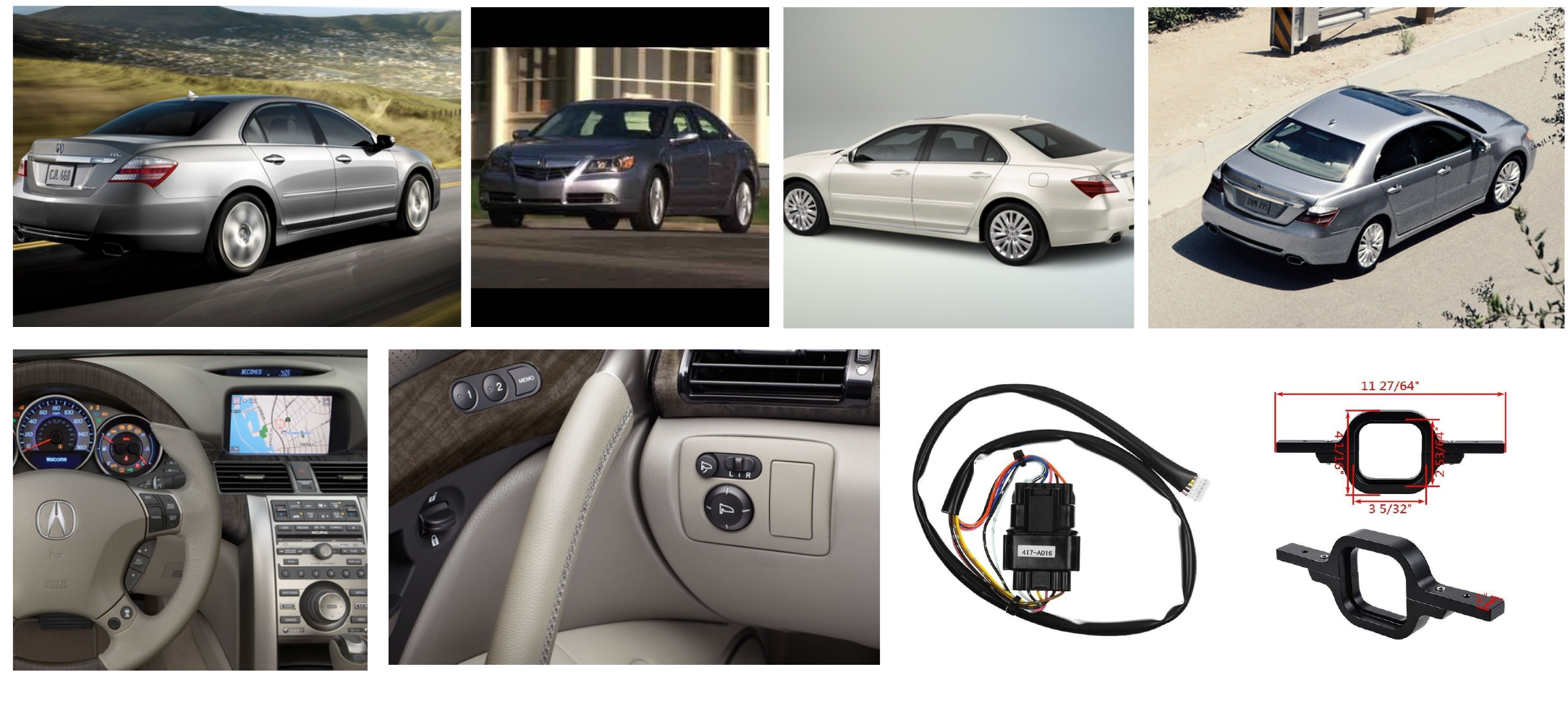}
	\caption{Example images in CARS-98N. Images in the first row have clean labels. The second row shows some images with noisy labels, including car interiors and car parts.}\label{fig:carsn_example}
\end{figure}

We compare the algorithms on five datasets, including:
\begin{itemize}
    \item \textbf{CARS} \cite{krause20133d}, which contains 16,185 images of 196 different car models. Following \cite{krause20133d}, we use the first 98 models for training and the rest for testing, and incorporate synthetic label noise into the training set.
    \item \textbf{CUB} \cite{wah2011caltech}, which contains 11,788 images of 200 different bird species. Following \cite{wah2011caltech}, we use the first 100 species for training and the rest for testing, and incorporate synthetic label noise into the training set.
    \item \textbf{Stanford Online Products (SOP)} \cite{oh2016deep}, which contains 59,551 images of 11,318 furniture items on eBay. We use 59,551 images in all classes for training and the rest for testing, and incorporate synthetic label noise into the training set.
    \item \textbf{Food-101N} \cite{lee2017cleannet}, which contains 310,009 images of food recipes in 101 classes. The test set is the Food-101 \cite{bossard14} dataset, which contains the same 101 classes as Food-101N. Images in Food-101N are obtained using search results from Google, Bing, Yelp, and TripAdvisor with an estimated noise rate of 20\% \cite{lee2017cleannet}. In the same evaluation setup with CARS, CUB, and SOP, we use 144,086 images in the first 50 classes (in alphabetical order) as the training set, and the remaining 51 classes in Food-101 as the test set which contains 51,000 images.
    \item \textbf{CARS-98N}. We build a new noisy label dataset named CARS-98N by crawling 9,558 images for 98 car models from Pinterest. We used the Pinterest image search engine to retrieve images using the 98 labels from the CARS training set as the query terms. The CARS-98N is \emph{only used for training}, and the test set of CARS is used for performance evaluation.  Figure~\ref{fig:carsn_example} shows example images in this dataset. The noisy images often contain the interior of the car, car parts, or images of other car models.
\end{itemize}

\subsection{Synthesizing Label Noise}
In the experiments, we adopt two models for noisy label synthesis: 1) symmetric noise and 2) Small Cluster noise. Symmetric noise \cite{van2015learning} has been widely used to evaluate the robustness of classification models (e.g., \cite{han2018co,yu2019does,patrini2017making,li2019learning}). Given a clean dataset, the symmetric noise model assigns a predefined portion of data from every ground-truth class to all other classes with equal probability, without regard to the similarity between data samples. After the noise synthesis, the number of classes remains unchanged. 


We contend that noisy labels that occur naturally differ from the symmetric noise model. Observing Food-101N and CARS-98N, the two datasets with naturally occurring noisy labels, we notice that some noisy data points are close to each other and can form small clusters. This is evident in Figure  \ref{fig:carsn_example}, where the car interior and car part images can form their own clusters. Further, the number of classes in metric learning may not be fixed. For example, in person or vehicle reidentification, two people or vehicles with similar looks may be inadvertently merged into one cluster. Conversely, images of the same person with different outfits may be separated into different clusters.  



To mimic these traits of naturally occurring label noise, we propose a new noise synthesis model --- Small Cluster. In this method, we first cluster images from a randomly selected ground-truth class into a large number of small clusters, using features extracted from a pretrained neural network. Here we use L2-normalized features from ResNet-18 pretrained on ImageNet. The number of clusters is set to one half of the number of images in the class. Each cluster is then merged into a randomly selected ground-truth class. After every iteration, the number of classes decreases by one. In this way, the Small Cluster model creates an open-set label noise scenario \cite{wang2018iterative} as the ground-truth classes are eliminated in the corrupted dataset. 

\begin{table*}[t!]
\centering
\caption{Precision@1 (\%) on CARS, SOP, and CUB dataset with symmetric label noise. 
}\label{tab:sym}
\resizebox{0.8\linewidth}{!}{ 
\begin{tabular}{@{}lrrrrrrrrrrr@{}}\toprule
     & \multicolumn{3}{c}{CARS}                &                   & \multicolumn{3}{c}{SOP}             &                       & \multicolumn{3}{c}{CUB}                                              \\ \cmidrule{2-4} \cmidrule{6-8} \cmidrule{10-12}
Noisy Label Rate & 10\%           & 20\%           & 50\% && 10\%           & 20\%           & 50\% && 10\%           & 20\%           & 50\%           \\ \midrule
\multicolumn{10}{l}{\textit{Algorithms for image classification under label noise}}                                                                                                                                 \\
Co-teaching \cite{han2018co}                      & 73.47          & 70.39          & 59.55        &             & 62.60          & 60.26          & 52.18               &      & 53.74          & 51.12          & 45.01          \\
Co-teaching+  \cite{yu2019does}                & 71.49          & 69.62          & 62.35         &            & 63.44          & 67.93          & 58.29              &       & 53.31          & 51.04          & 45.16          \\

Co-teaching \cite{han2018co} w/ Temperature \cite{zhaiclassification}           & 77.51          & 76.30          & 66.87      &               & 73.71          & 71.97          & 64.07           &         & 55.25          & 54.18          & 50.65          \\ F-correction   \cite{patrini2017making}                   & 71.00          & 69.47          & 59.54     &                & 51.18          & 46.34          & 48.92               &      & 53.41          & 52.60          & 48.84          \\ 
\midrule
\multicolumn{10}{l}{\textit{DML with Proxy-based Losses}}                                                                                                                                                                             \\
FastAP      \cite{cakir2019deep}                      & 66.74          & 66.39          & 58.87         &            & 69.20          & 67.94          & 65.83              &       & 54.10          & 53.70          & 51.18          \\
nSoftmax     \cite{zhaiclassification}                     & 72.72          & 70.10          & 54.80         &            & 70.10          & 68.90          & 57.32          &           & 51.99          & 49.66          & 42.81          \\
ProxyNCA     \cite{movshovitz2017no}                     & 69.79          & 70.31          & 61.75           &          & 71.10          & 69.50          & 61.49         &            & 47.13          & 46.64          & 41.63          \\
Soft Triple    \cite{qian2019softtriple}                    & 76.18          & 71.82          & 52.53        &             & 68.60          & 55.21          & 38.45            &         & 51.94          & 49.14          & 41.46          \\
\midrule
\multicolumn{10}{l}{\textit{DML with Pair-based Losses}}                                                                                                                                                                             \\
MS        \cite{wang2019multi}                        & 66.31          & 67.14          & 38.24    &  & 69.90          & 67.60          & 59.58       &              & 57.44          & 54.52          & 40.70          \\
Circle     \cite{sun2020circle}                       & 71.00          & 56.24          & 15.24         &            & 72.80          & 70.50          & 41.17          &           & 47.48          & 45.32          & 12.98          \\
Contrastive Loss  \cite{chopra2005learning}                     & 72.34          & 70.93          & 22.91          &           & 68.70          & 68.80          & 61.16       &              & 51.77          & 51.50          & 38.59          \\
Memory Contrastive Loss (MCL)  \cite{wang2020cross}                & 74.22          & 69.17          & 46.88       &              & 79.00          & 76.60          & 67.21               &      & 56.72          & 50.74          & 31.18          \\
\midrule
MCL + PRISM (Ours)    & \textbf{80.06} & \textbf{78.03} & \textbf{72.93}    &        & \textbf{80.11} & \textbf{79.47} & \textbf{72.85}       &     & \textbf{58.78}&  \textbf{58.73} & \textbf{56.03} \\ \bottomrule
\end{tabular}
}
\end{table*}

\subsection{Baseline Techniques}\label{sec:setting}
We compare PRISM against 12 baselines, including four baselines designed for noise-resistant classification and eight baselines for deep metric learning:
\begin{itemize}
    \item Co-teaching \cite{han2018co}, which trains two CNNs jointly. Data samples assigned small losses by one model are used to train the other model. 
    \item Co-teaching+ \cite{yu2019does}. Similar to Co-teaching, a model is trained using data samples that (1) the two models disagree on, and (2) receive small losses from the other model.
    \item Co-teaching \cite{han2018co} with Temperature \cite{zhaiclassification}, which adds a temperature hyperparameter to the cross-entropy loss.
    \item F-correction \cite{patrini2017making}, which multiplies the class transition matrix to the loss function. Since Small Cluster produces open-set label noises, for which the class transition matrix is not properly defined, this baseline is only used under symmetric noise settings.
    \item Eight DML algorithms. Four of them use proxy-based losses (including Soft Triple \cite{qian2019softtriple}, FastAp \cite{cakir2019deep}, nSoftmax \cite{zhaiclassification} and proxyNCA \cite{movshovitz2017no}), and the other four use pair-based losses (including MS loss \cite{wang2019multi}, circle loss \cite{sun2020circle}, contrastive loss \cite{chopra2005learning} and memory contrastive loss (MCL) \cite{wang2020cross}). They treat all data samples as clean.
\end{itemize}
We train the classification baselines using cross-entropy. When retrieving images during inference, we use the L2-normalized features from the layer before the final linear classifier.

For Co-teaching, Co-teaching+ and Memory Contrastive Loss \cite{wang2020cross}, we use the official implementation. The other DML algorithms are implemented by Pytorch Metric Learning \cite{musgrave2020pytorch}. 
We follow the hyperparameter settings given in the respective papers or code repositories. For Co-teaching and Co-teaching+, we use the learning rate (LR) scheduler given in their code, while for others (including our algorithm), cosine LR decay \cite{loshchilov2016sgdr} is used. We set the batch size to 64 for experiments on all datasets and all models. The input images are first resized to 256x256, then randomly cropped to 224x224. A horizontal flip is performed on the training data with a possibility of 0.5. 

For Soft Triple and Soft Triple with PRISM, we set the number of proxies per class $H=10$ for CARS-98N and Food-101N. Other hyperparameters follow \cite{qian2019softtriple}. The size of the memory bank is set to the size of the training dataset as in \cite{wang2020cross}.

Again following \cite{wang2020cross}, when comparing performance on CARS, CUB and CARS-98N, we use BN-inception \cite{ioffe2015batch} as the backbone CNN model for all algorithms. The dimension of the output feature is set as 512, the same as in \cite{wang2020cross}. No other tricks (e.g., freezing BN layers) are used during the experiments. For SOP and Food-101N, we use ResNet-50 \cite{he2016deep} with a 128-dimensional output. 

Testing is based on the ranked list of the nearest neighbors for the test images. Specifically, we use Precision@1 (P@1) and Mean Average Precision@R (MAP@R) \cite{musgrave2020metric} as the evaluation metrics. The test sets are noise-free, as the purpose of the experiments is to evaluate the algorithms in the presence of noisy labels in the training data.

\subsection{Results and Discussions}

\begin{table}[t!]
\centering
\caption{Precision@1 (\%) on CARS, SOP, and CUB with Small Cluster label noise. 
}\label{tab:splitmerge}
\resizebox{1\columnwidth}{!}{ 
\begin{tabular}{@{}lrrrrrrrr@{}}\toprule
     & \multicolumn{2}{c}{CARS}             &      & \multicolumn{2}{c}{SOP}             &        & \multicolumn{2}{c}{CUB}                     \\ 
     \cmidrule{2-3} \cmidrule{5-6} \cmidrule{8-9}
Noisy Label Rate & 25\%           & 50\%&& 25\%&50\%& &25\%                  & 50\%         \\ \midrule
\multicolumn{9}{l}{\textit{Algorithms for image classification under label noise}}                                                                               \\
Co-teaching                       & 70.57          & 62.91            &        & 61.97               & 58.08         &           & 51.75                     & 48.85          \\
Co-teaching+                  &    70.05         & 61.58          &          & 62.57               & 59.27       &             & 51.55                       & 47.62          \\                                                       
Co-teaching w/ Temperature            & 75.26          & 66.19        &            & 70.19               & 68.50      &              & 54.59                     & 48.32          \\ 
\midrule
\multicolumn{9}{l}{\textit{DML with Proxy-based Losses}}                                                                                                                            \\
FastAP                            & 62.49          & 53.07       &             & 70.66               & 67.55      &              & 52.18                     & 48.46          \\
nSoftmax                          & 71.61          & 62.29       &             & 70.00               & 61.92       &             & 49.61                     & 41.78          \\
ProxyNCA                          & 69.50          & 58.34       &             & 67.95               & 62.25         &           & 42.07                     & 36.48          \\ 
Soft Triple                        & 73.26          & 66.66       &             & {\bf 73.63}               & 64.14         &           & 56.18                     & 50.35          \\
Soft Triple + PRISM (Ours)      & {\bf 77.60}         & {\bf 70.45}         &               & 70.99       & {\bf 69.38}       &           & {\bf 57.61}            & {\bf 54.27} \\
\midrule
\multicolumn{9}{l}{\textit{DML with Pair-based Losses}}                                                     \\
MS                                & 63.92          & 43.73          &          & 67.32               & 62.17        &            & 53.60                     & 41.66          \\
Circle                            & 53.03          & 19.95          &          & 70.33               & 40.48           &         & 44.07                     & 22.96          \\
Contrastive Loss                      & 65.60          & 26.45         &           & 68.25               & 64.27          &          & 47.27                     & 39.43          \\
Memory Contrastive Loss (MCL)                & 69.46          & 36.43           &         &  75.61              & 68.71       &             & 52.25                     & 41.58          \\ 
MCL + PRISM (Ours)    & {\bf 77.08}          & {\bf 68.26}         &           & {\bf 78.56}      & {\bf 73.84}     &      & {\bf 55.77}                     & {\bf 53.46}          \\
 \bottomrule
\end{tabular}
}

\end{table}

\begin{table}[t!]
\centering
\caption{Precision@1 (\%) and Mean Average Precision@R (\%) on CARS-98N and Food-101N~\cite{lee2017cleannet}.}\label{tab:cars98n}
\resizebox{1\linewidth}{!}{\begin{tabular}{@{}lrrrrr@{}}\toprule
                               & \multicolumn{2}{c}{CARS-98N}   &   & \multicolumn{2}{c}{Food-101N}       \\ 
                               \cmidrule{2-3} \cmidrule{5-6}
                               & P@1             & MAP@R      &     & P@1            & MAP@R          \\ \midrule
\multicolumn{6}{l}{\textit{Algorithms for image classification under label noise}} \\
Co-teaching                    & 58.74           & 9.10     &       & 59.08          & 14.66          \\
Co-teaching+               & 56.66           & 8.40     &       & 57.59          & 14.72          \\
Co-teaching w/ Temperature   & 60.72           & 9.61      &      & 63.18          & 17.38          \\ 
\midrule
\multicolumn{6}{l}{\textit{DML with Proxy-based Losses}}                                               \\
ProxyNCA                       & 53.55           & 8.75      &      & 48.41          & 9.30           \\ 
Soft Triple                    & 63.36           & 10.88     &       & 63.61          & 16.23          \\
Soft Triple + PRISM (Ours)      & {\bf 64.81}         & {\bf 11.21}     &       & {\bf 64.46}         & {\bf 17.53} \\
\midrule
\multicolumn{6}{l}{\textit{DML with Pair-based Losses}}                                              \\
MS                             & 49.00           & 5.92     &       & 52.53          & 9.82           \\
Contrastive                    & 44.91           & 4.76     &       & 50.04          & 9.42           \\
Memory Contrastive Loss (MCL)              & 38.73           & 3.34      &      & {\bf 52.58}          & {\bf 9.88}           \\ 
MCL + PRISM (Ours)  & {\bf 57.95}           & {\bf 8.04}     &       & 52.47          & 9.64               \\
 \bottomrule
\end{tabular}}
\end{table}

\noindent \textbf{Symmetric Label Noise. } Table~\ref{tab:sym} shows the evaluation results on CARS, SOP, and CUB under symmetric label noise. 
PRISM with Memory Contrastive Loss achieves the highest performance among all the compared algorithms. As the noisy label rate increases, the Precision@1 scores decrease for all approaches, but PRISM exhibits the least performance drop among all methods. 
In CUB, the performance of PRISM decreases by 0.05\% when the noise level increases from 10\% to 20\%, and by less than 3\% when noise increases to 50\%. 

The robust classification methods, including Co-teaching, Co-teaching+, and F-correction, show a certain level of robustness to noisy labels. However, their performance is generally lower compared to DML approaches and PRISM. Applying temperature normalization to cross-entropy \cite{zhaiclassification} boosts the performance of Co-teaching, especially on the SOP dataset.

Approaches with proxy-based loss achieve generally high scores under CUB and CARS, even at 50\% noise. However, they generally perform worse than approaches with pair-based loss under SOP. The average number of images per class in the SOP training set is only 5.26, in contrast to 82.18 in CARS, causing difficulties in accurately estimating the proxy centers in SOP. 

In comparison, pair-based losses are subject to severe performance drops under high noise rates. The pair-based loss takes a pair of samples $(x_i,x_j)$ as unit. At 50\% noise, the correct pairs only account for 25\% of all pairs, which bears on the performance heavily. In addition, both MS loss \cite{wang2019multi} and circle loss \cite{sun2020circle} assign higher weights for learning the hard examples which could be the noisy data. This causes the performance to drop significantly. The use of the memory bank in MCL increases the performance. However, the performance is still lower than that of proxy-based loss approaches.


\vspace{0.05in}
\noindent  \textbf{Small Cluster Open-set Label Noise.} Table~\ref{tab:splitmerge} reports the Precision@1 scores achieved by various approaches on datasets with Small Cluster label noise. Incorporating PRISM into MCL improves the performance of the resulting model. The advantage of PRISM is especially pronounced under 50\% noise rate. A similar trend of performance can be observed for SoftTriple + PRISM.

\vspace{0.05in}
\noindent \textbf{Real-world Noisy Datasets.} Table~\ref{tab:cars98n} displays the performance on two datasets with real-world label noise, CARS-98N and Food-101N~\cite{lee2017cleannet}. 
In both datasets, using Soft Triple with PRISM achieves the best performance. 
In CARS-98N, MCL performs worse than the na\"ive contrastive loss \cite{chopra2005learning} because the presence of noise in the memory bank hurts the performance of MCL. However, after adding PRISM to MCL, we improved Precision@1 by as much as 19\%. However, we do not observe performance improvement on Food-101N from incorporating PRISM into MCL. We attribute this to an interesting characteristic of Food-101N, that many small clusters of open-set noise are often unique to one ground-truth class. For example, images of oatmeals often appear in the class \texttt{apple-pie}, but not in similar classes like \texttt{crab-cakes} or \texttt{chocolate-cake}. As a result, the model may have learned to treat oatmeals to be special apple pies. The multi-center SoftTriple loss is less susceptible to this phenomenon. 


\begin{table}[t]
\caption{Training time required with and without PRISM for 5,000 iterations on the SOP dataset and 10\% symmetric label noise. The time recording starts when the memory bank is completely filled at iteration 3,000.}\label{tab:time}
\centering
\resizebox{1\columnwidth}{!}{ 
\begin{tabular}{@{}lr@{}}\toprule
Algorithm              & Training Time (Seconds) \\ \midrule
Memory Contrastive Loss (MCL)       & 1,679.22        \\ 
MCL + PRISM without centers         & 12,294.76       \\ 
MCL + PRISM with centers            & 1,777.38        \\ 
\midrule
Soft Triple                         & 1,685.47        \\ 
Soft Triple + PRISM with centers    & 1,767.97        \\ \bottomrule
\end{tabular}
}
\end{table}

\vspace{0.05in}
\noindent \textbf{Training Time.} Table~\ref{tab:time} reports the time required for 5,000 training iterations. PRISM without centers (Eq.~\eqref{equ:pi}) requires the longest training time because it needs to calculate average similarities of all classes for each minibatch data. However, by maintaining center vectors to identify noise (Eq.~\eqref{equ:pi2}), the required training time decreases significantly. PRISM only incurs 6\% more training time than that of Memory Contrastive Loss on SOP Similar observation can be found when we change the loss function to Soft Triple \cite{qian2019softtriple}. On CARS and CUB dataset, PRISM incurs up to 10.4\% more training time. Due to space limitation, we show detailed results in the supplementary material. The results show that PRISM is an efficient method to handle noisy labels.

\subsection{Ablation Study}
\noindent \textbf{Design of $P_{\text{clean}}(i)$.} We compare PRISM against slightly different designs of the clean-data identification function $P_{\text{clean}}(i)$. As baselines, we adopt the following designs of $P_{\text{clean}}(i)$, which are compared with Eq.~\eqref{equ:pi}. The first baseline, Batch-positive, uses the average similarity between $x_i$ and same-class samples within the minibatch. This approach neither uses the memory bank nor considers the negative samples.
\begin{equation}\label{equ:ablation_design_noMemory}
    P_{\text{clean}}(i) \propto \frac{1}{M_{y_i}}\sum_{(x_j,y_j) \in \mathcal{B}, y_j=y_i } S(f(x_i),f(x_j))
\end{equation}
The second baseline, Memory-positive, uses the average similarity between $x_i$ and other same-class samples retrieved from the memory bank, but still does not consider the negative samples.
\begin{equation}\label{equ:ablation_design_withMemory}
    P_{\text{clean}}(i) \propto \frac{1}{M_{y_i}}\sum_{(v_j,y_j) \in \mathcal{M}, \,y_j=y_i}S(f(x_i),v_j)
\end{equation}

Table~\ref{tab:pi_compare} reports the performance on CARS with 25\% Small Cluster noise using different noise-filtering strategies with the threshold $m$ determined by TRM. We use Soft Triple \cite{qian2019softtriple} as the loss function. All filtering strategies improve performance relative to the scenario of no noise filtering. The Memory-positive strategy works better than Batch-positive, showing the importance of using the entire memory bank to identify noise. PRISM, in the form of Equation~\eqref{equ:pi}, achieves the best P@1 and MAP@R scores. 

\begin{figure}[t]
  \centering
  \includegraphics[width=1\columnwidth]{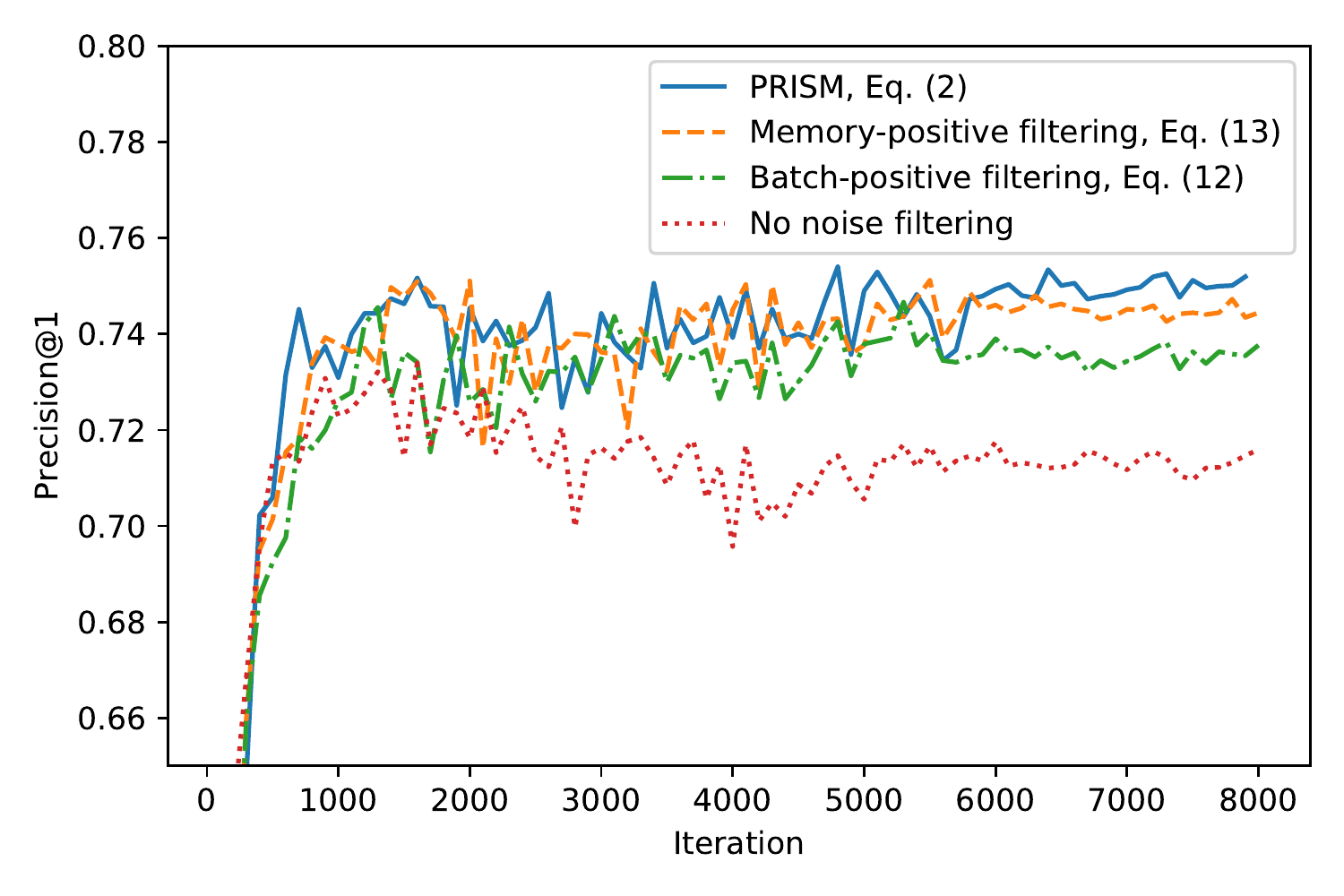}
  \caption{The Precision@1 (\%) vs number of iterations on CARS with 25\% Small Cluster Noise.}\label{fig:piplot}
\end{figure}

Figure~\ref{fig:piplot} shows that the Precision@1 changes as training iterations increase. Ignoring label noise makes the model overfit to the noisy data, which is shown as good initial performance followed by a rapid fall. The batch-positive strategy, which does not use the memory bank, can improve performance, but also experiences a performance drop after 5,000 training iterations. Under the the PRISM strategy of Eq.~\eqref{equ:pi}, the model converges to the best performance.

\begin{table}[t]
\centering
\caption{Precision@1 (\%) and MAP@R (\%) under different designs of the noise identification function. We use TRM to determine the threshold $m$ for all methods. 
}\label{tab:pi_compare}
\begin{tabular}{@{}lrr@{}}\toprule
                                         & P@1   & MAP@R \\ \midrule
No noise filtering            & 73.40 & 15.21 \\ 
Batch-positive filtering, Eq.~\eqref{equ:ablation_design_noMemory} & 74.66 & 16.44 \\ 
Memory-positive filtering, Eq.~\eqref{equ:ablation_design_withMemory}  & 75.12 & 17.11 \\ 
PRISM, Eq.~\eqref{equ:pi}  & \textbf{75.40} & \textbf{17.40} \\ \bottomrule
\end{tabular}
\end{table}

\begin{figure}[t]
  \centering
  \includegraphics[width=1\columnwidth]{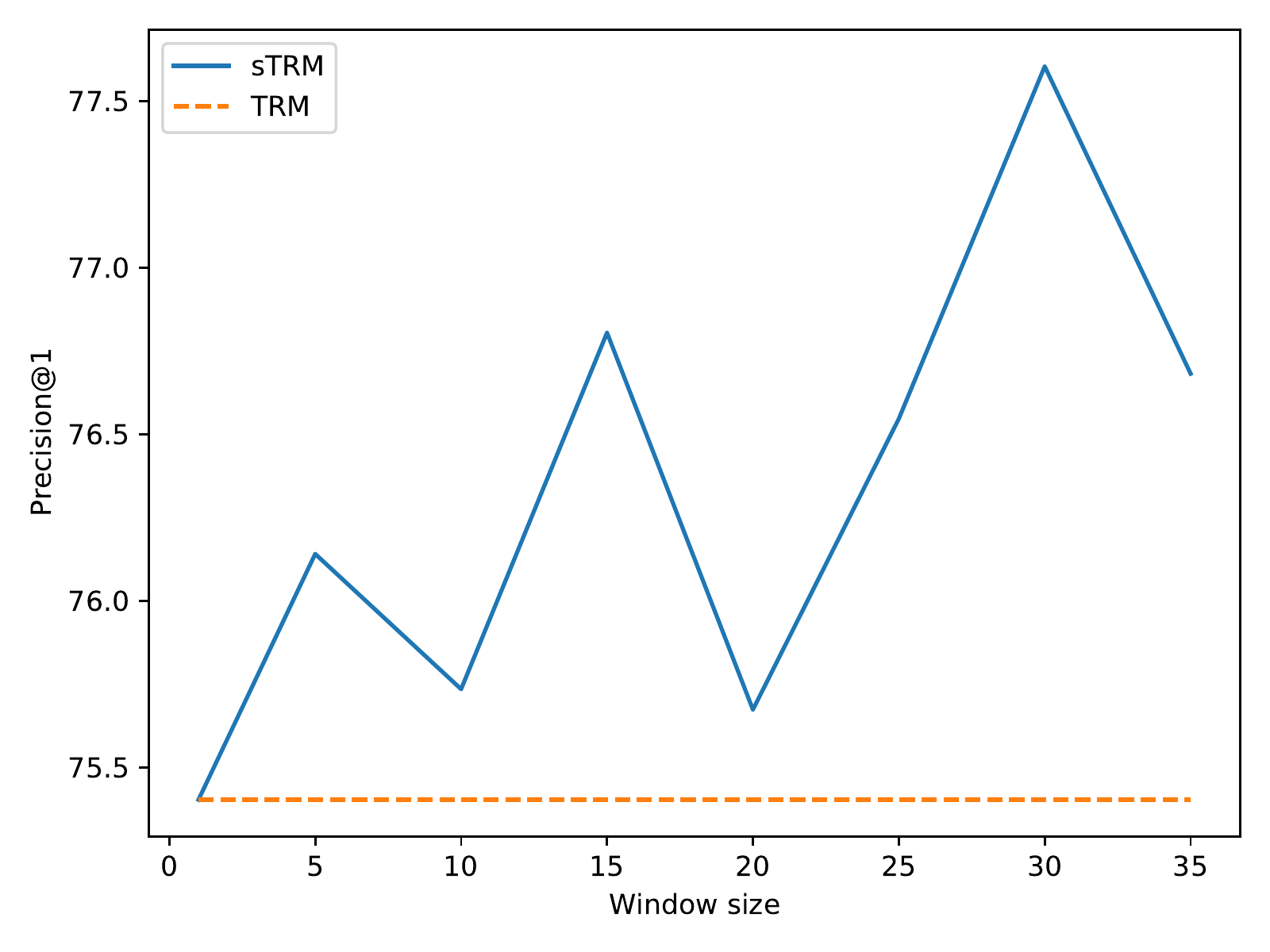}
  \caption{The Precision@1 (\%) vs. window size $\tau$. The dataset used is CARS with 25\% Small Cluster noise. 
  }\label{fig:trm}
\end{figure}

\vspace{0.05in}
\noindent \textbf{TRM vs sTRM.} Figure~\ref{fig:trm} illustrates the model performance for different choices of the sliding window size in sTRM. Note that TRM is a special case of sTRM, where the window size $\tau$ is set to $1$. Across all different choices of $\tau$, sTRM consistently outperforms TRM.

\vspace{0.05in}
\noindent \textbf{Noise rate $R$ for CARS-98N.} In PRISM, we use the $R^{\text{th}}$ percentile of $P_{\text{clean}}(i)$ values to determine the threshold $m$ for identifying noisy labels. Table~\ref{tab:R_compare} shows the performance under different $R$ values. We use memory-based contrastive loss \cite{wang2020cross} as the loss function. The model achieves the best performance when $R=50\%$. We can thus estimate that the noisy label rate of CARS-98N is approximately $50\%$. We also use $R=50\%$ for Co-teaching and its extensions.

\begin{table}[t!]
\caption{Precision@1 (\%) and Mean Average Precision@R (\%) when using different filtering rate $R$ for identifying noisy label. Models trained with filtering rate $R=50\%$ obtained the best performance.}\label{tab:R_compare}
	\resizebox{1\linewidth}{!}{\begin{tabular}{@{}lrrrrrr@{}}
\toprule
$R$   & 0.0   & 0.3   & 0.4   & 0.5   & 0.6   & 0.7   \\ \midrule
P@1   & 38.73 & 49.43 & 52.43 & \textbf{57.95} & 54.91 & 51.27 \\ 
MAP@R & 3.34  & 5.47  & 6.29  & \textbf{8.04}  & 7.43  & 6.29  \\ \bottomrule
\end{tabular}}
\end{table}

\section{Conclusions}
In this paper, we propose a simple, efficient, and effective approach, Probabilistic Ranking-based Instance Selection with Memory (PRISM), to enhance the performance of deep metric learning in the presence of training label noise. Through extensive experiments with both synthetic and real-world datasets, we demonstrate that PRISM outperforms 12 existing approaches. 

\section{Acknowledgments}
We gratefully acknowledge the support by the National Research Foundation, Singapore through the AI Singapore Programme (AISG2-RP-2020-019), NRF Investigatorship (NRF-NRFI05-2019-0002), and NRF Fellowship (NRF-NRFF13-2021-0006); Alibaba Group through Alibaba Innovative Research and Alibaba-NTU Singapore Joint Research Institute (Alibaba-NTU-AIR2019B1); the Nanyang Assistant/Associate Professorships; NTU-SDU-CFAIR (NSC-2019-011); NSFC No.91846205; the Innovation Method Fund of China No.2018IM020200; the RIE 2020 Advanced Manufacturing and Engineering Programmatic Fund (No. A20G8b0102), Singapore. 


{\small
\bibliographystyle{ieee_fullname}
\bibliography{main}
}
\end{document}


\title{Supplementary Materials for \\\ Noise-resistant Deep Metric Learning with Ranking-based Instance Selection}

\author{}

\maketitle

\section{Generating Small Cluster Noise}

To mimic characteristics of natural label noise, we propose a new model of noise synthesis called Small Cluster. In Algorithm~\ref{alg:noise}, we show the pseudo-code for generating Small Cluster noise from a clean dataset. 
The algorithm first clusters images from a randomly selected ground-truth class into a large number of small clusters, using features extracted from a pretrained neural network. The number of clusters is set to $1/Z$ of the number of images in the class so each cluster is expected to have $Z$ images. Each cluster is then merged into a randomly selected ground-truth class. The procedure is repeated until, out of the total of $N$ images, the number of misplaced images reaches or exceeds the predefined percentage $R$. 

In our experiments, we choose $Z = 2$ and set the random seed to 0. We use Mini-batch K-means \cite{sculley2010web} as our clustering algorithm.

\begin{algorithm}[h!]
	\SetAlgoLined
	\Input{$\mathcal{X}=\{(x_0,\Tilde{y_0}),(x_1,\Tilde{y_1}),...,(x_N,\Tilde{y_N})\}$: training dataset\\$R$: the noise rate \\ $Z$: mean number of images per cluster}
	\Output{$Y$: the corrupted labels}
	\BlankLine
	$Y=[\Tilde{y_0},\Tilde{y_1},...,\Tilde{y_N}]$\\
    \While{\#misplaced\_images $<$ RN}{
        $c$ = a uniformly sampled non-empty class\\
        $X_{c} = \{x_i| (x_i,\Tilde{y_i}) \in \mathcal{X}, \Tilde{y_i}=c \}$\\
        $Q=$ Clustering($X_{c}$,n\_cluster=int($\frac{|X_{c}|}{Z}$))\\
        \For{$q_i \in Q$}{
            $c'$ = another uniformly sampled non-empty class that does not equal $c$\\
            \For{$x_j \in q_i$}{
                $Y[j]=c'$
            }
        }
        Mark $\forall x \in X_{c}$ as misplaced images
    }
    \Return{Y}
	\BlankLine
	\caption{Synthesizing Small Cluster Noise}\label{alg:noise}
\end{algorithm}

\section{Training time on CARS and CUB.} Table~\ref{tab:time_appendix} shows the training time with and without PRISM algorithm. PRISM adds about 100 seconds for 5K iterations, or 8\% to 10\% of total running time, on CARS and CUB.
\begin{table}[h!]
\centering
\resizebox{1\columnwidth}{!}{
\begin{tabular}{@{}lll@{}}\toprule
Algorithm                           &  CARS                             &  CUB          \\ \midrule
MCL                                 & 1,170.05                          & 1,189.91       \\ 
MCL + PRISM                         & 1,291.58 (+10.4\%)                & 1,305.90 (+8.0\%)      \\ 
\midrule
Soft Triple                         & 1,186.658                         & 1,184.75       \\ 
Soft Triple + PRISM                 & 1,279.97 (+7.9\%)                 & 1,284.97 (+10.2\%)      \\ \bottomrule
\end{tabular}}
\caption{Training time (seconds) for 5K iterations.}\label{tab:time_appendix}
\end{table}
\section{Results on Landmark Recognition}
We conduct experiments on landmark recognition datasets. We use Oxford \cite{philbin2007object} and Babenko's Landmark dataset \cite{babenko2014neural} to train our model. RParis \cite{radenovic2018revisiting} is used to test the performance. Details of the dataset are described below.
\begin{itemize}
    \item The \textbf{Oxford} Dataset \cite{philbin2007object} consists of 5,062 images of 11 Oxford landmarks, collected from Flickr. We utilize all the images (including images in which the buildings are not present, heavily occluded, or distorted).
    \item \textbf{Babenko's Landmark} Dataset \cite{babenko2014neural} consists of 213,678 images of 672 landmarks. The images are retrieved by querying the Yandex image search engine with the name of landmarks. Certain level of label noise exists \cite{gordo2017end}.
    \item The Revisited Paris (\textbf{RParis}) Dataset \cite{radenovic2018revisiting} contains 6,412 images of 12  landmarks in Paris. The dataset is originally created by \cite{philbin2008lost} then cleaned by \cite{radenovic2018revisiting}.
\end{itemize}
The training setting follows Section 4.3 of the main paper.

Results show that PRISM improves the performance on both small- and large-scale landmark recognition datasets with significant levels of label noise. 
\begin{table}[h!]
\caption{mAP on RParis. Models are trained on Oxford Dataset.}\label{tab:oxford}
\centering
\begin{tabular}{@{}lrrr@{}}\toprule
Alorithm        & Easy      & Medium    & Hard      \\ \midrule
MCL             & 60.8      & 47.9     & 24.8       \\ 
MCL+PRISM       & \textbf{61.7}      & \textbf{48.8}     & \textbf{25.7}       \\ 
\midrule
Soft Triple             & 63.9      & 49.9     & 25.2       \\ 
Soft Triple+PRISM       & \textbf{64.1}      & \textbf{50.1}     & \textbf{26.5}       \\ 
\midrule
\end{tabular}
\end{table}

\begin{table}[h!]
\caption{Precision@1 and MAP@R on RParis. Models are trained on Babenko's Landmark Dataset.}\label{tab:babenko}
\centering
\begin{tabular}{@{}lrr@{}}\toprule
                                & P@1             & MAP@R \\ \midrule
MCL   & 82.04           & 21.80 \\ 
MCL + PRISM (Ours)              & {\bf 82.98}     & {\bf 22.33} \\
 \bottomrule
\end{tabular}
\end{table}
\section{Results on Clean Datasets}
In Table~\ref{tab:nonoise}, we report the results when the algorithm is trained on the original CUB, CARS, and SOP datasets. The training setting is identical to that in Section 4.3 of the main paper. The performance degradation on CUB is small. On SOP, filtering data at $R=2\%$ and $5\%$ causes performance to improve slightly. After inspection, we believe the original SOP dataset contains some noisy labels, indicating that noisy labels are common in real-world data. 

\begin{table}[h!]
\caption{Precision@1 under different filtering rate $R$ for MCL with PRISM.}\label{tab:nonoise}
\centering
\begin{tabular}{@{}lrrrr@{}}\toprule
Dataset   & MCL only (R=0)  &R=2\% &R=5\% &R=10\%      \\ \midrule
CUB     & 60.8    & 60.4    & 60.0  & 60.1      \\ 
CARS    & 82.1    & 81.3    & 80.2  & 79.3   \\ 
SOP     & 81.0    & 81.2    & 81.1  & 80.8     \\ 
\midrule
\end{tabular}
\end{table}

\section{Details of CARS-98N}
Using the 98 labels from the CARS training set as the query terms, We build the CARS-98N dataset from the image search of Pinterest. No data cleaning has been performed. CARS-98N is only used for training. The size of CARS-98N is about $20\%$ times larger than the training set of CARS \cite{krause20133d} dataset.

\begin{figure*}[b]
      \centering
      \includegraphics[width=1\linewidth]{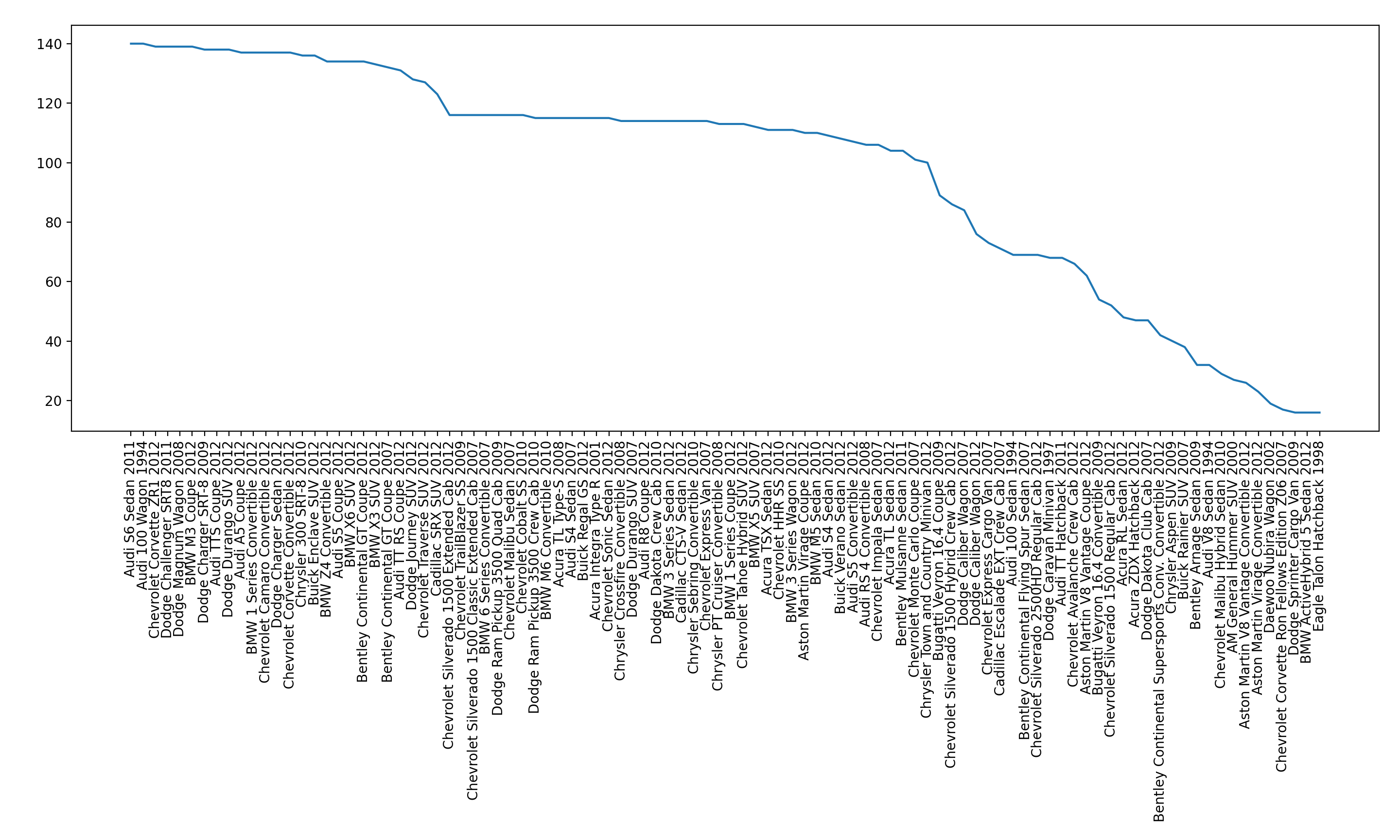}
	\caption{The number of images for each class in CARS-98N dataset. The X-axis gives the car model name and Y-axis refers to the number of images. }\label{fig:carsn_overview}
\end{figure*}

Figure~\ref{fig:carsn_overview} shows the number of images for each class in CARS-98N dataset. It can be observed that the number of images is not evenly distributed across classes. Although many classes contain more than 100 images, fewer images can be found for certain car models such as Chevrolet Malibu Hybrid 2010, probably due to their limited market share or availability.
\begin{figure*}[h!]
    \begin{subfigure}{0.23\linewidth}
      \centering
      \includegraphics[width=1\columnwidth]{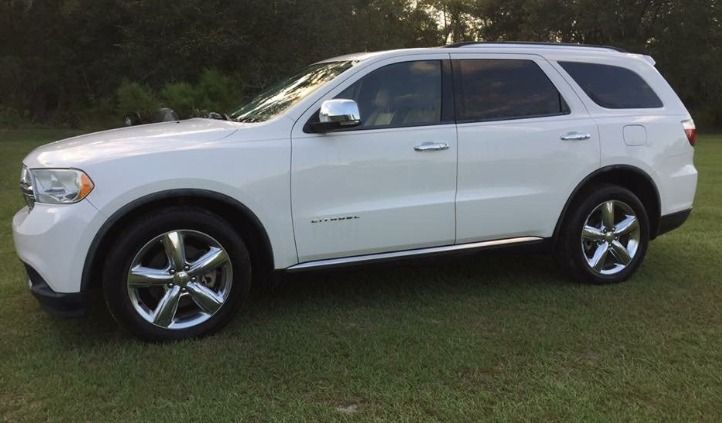}
    \end{subfigure}
    \begin{subfigure}{0.23\linewidth}
      \centering
      \includegraphics[width=1\columnwidth]{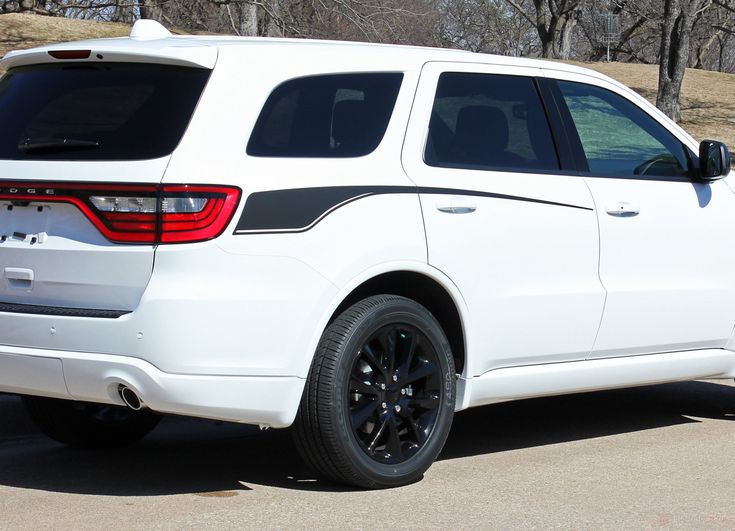}
    \end{subfigure}		
        \begin{subfigure}{0.23\linewidth}
      \centering
      \includegraphics[width=1\columnwidth]{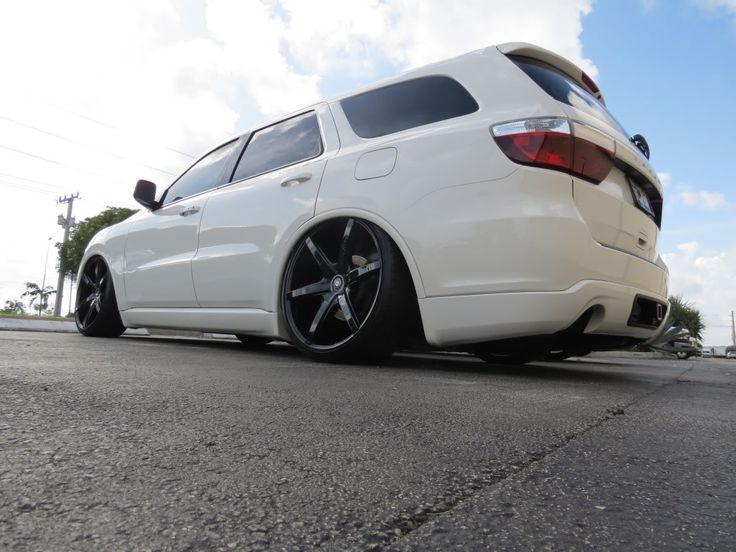}
    \end{subfigure}	
        \begin{subfigure}{0.23\linewidth}
      \centering
      \includegraphics[width=1\columnwidth]{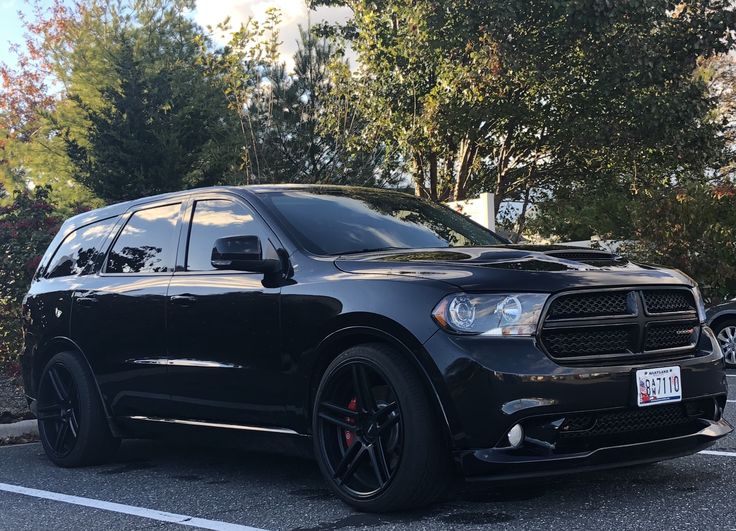}
    \end{subfigure}	
	\caption{Images of Dodge Durango SUV 2012 that are correctly labeled in CARS-98N dataset.}\label{fig:correct}
\end{figure*}
\begin{figure*}[h!]
    \begin{subfigure}{0.23\linewidth}
      \centering
      \includegraphics[width=1\columnwidth]{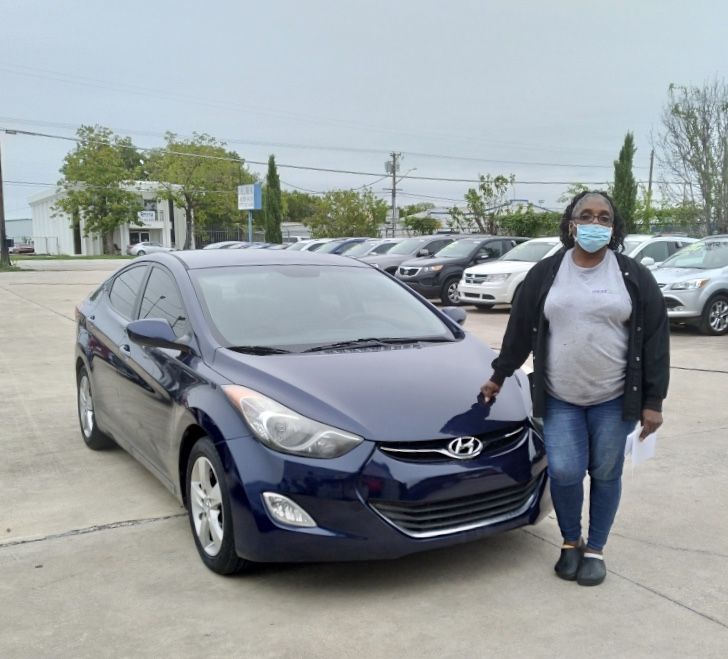}
    \end{subfigure}
    \begin{subfigure}{0.23\linewidth}
      \centering
      \includegraphics[width=1\columnwidth]{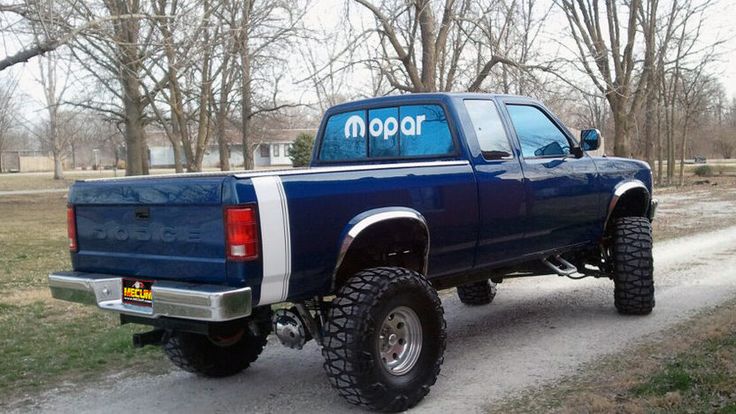}
    \end{subfigure}		
        \begin{subfigure}{0.23\linewidth}
      \centering
      \includegraphics[width=1\columnwidth]{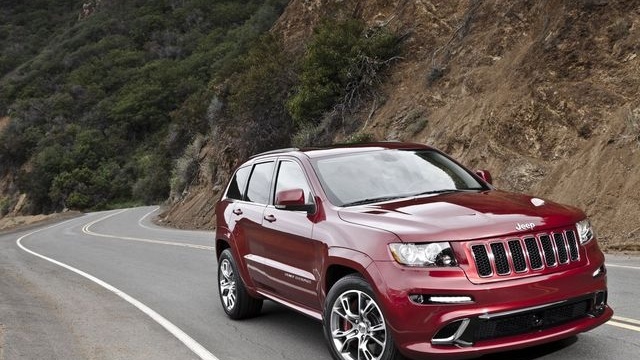}
    \end{subfigure}	
        \begin{subfigure}{0.23\linewidth}
      \centering
      \includegraphics[width=1\columnwidth]{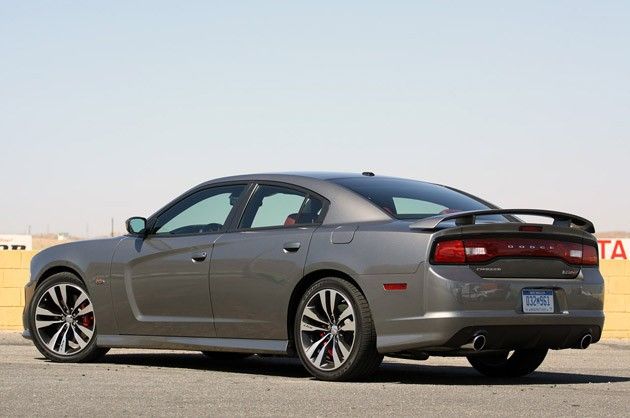}
    \end{subfigure}	
	\caption{Incorrect car models found in the class Dodge Durango SUV 2012 in CARS-98N.}\label{fig:othercar}
\end{figure*}
\begin{figure*}[h!]
    \begin{subfigure}{0.23\linewidth}
      \centering
      \includegraphics[width=1\columnwidth]{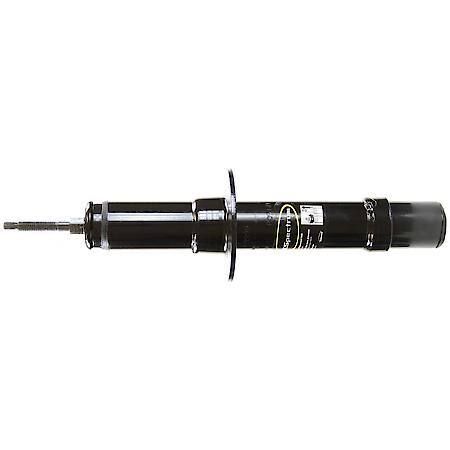}
    \end{subfigure}
    \begin{subfigure}{0.23\linewidth}
      \centering
      \includegraphics[width=1\columnwidth]{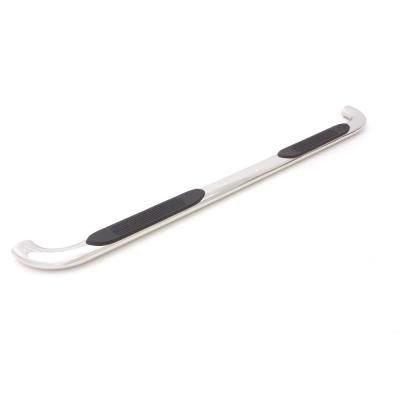}
    \end{subfigure}		
        \begin{subfigure}{0.23\linewidth}
      \centering
      \includegraphics[width=1\columnwidth]{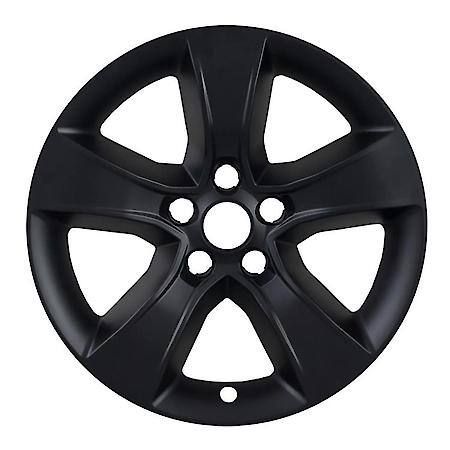}
    \end{subfigure}	
        \begin{subfigure}{0.23\linewidth}
      \centering
      \includegraphics[width=1\columnwidth]{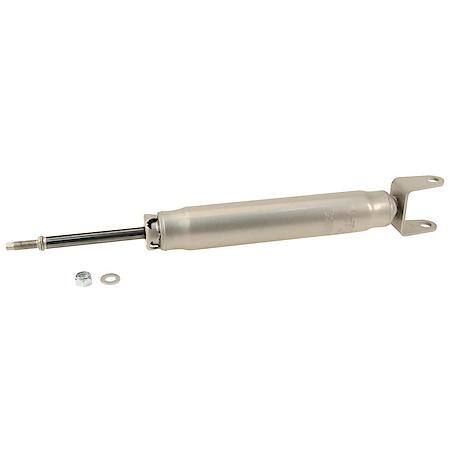}
    \end{subfigure}	
	\caption{Car part and accessory images in the class Dodge Durango SUV 2012 in the CARS-98N dataset.}\label{fig:parts}
\end{figure*}
\begin{figure*}[h!]
    \begin{subfigure}{0.23\linewidth}
      \centering
      \includegraphics[width=1\columnwidth]{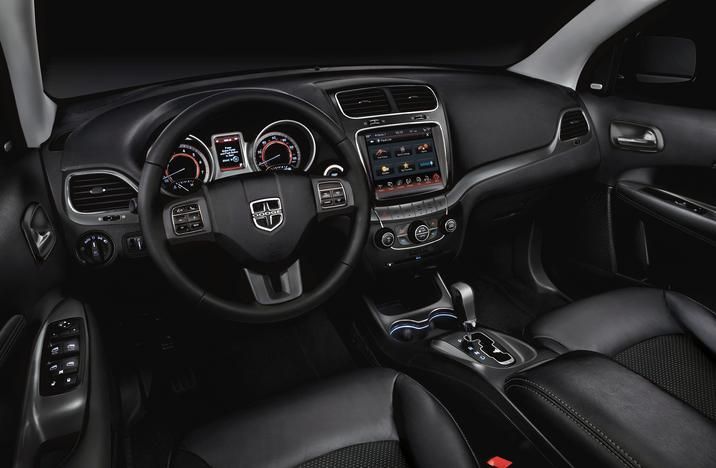}
    \end{subfigure}
    \begin{subfigure}{0.23\linewidth}
      \centering
      \includegraphics[width=1\columnwidth]{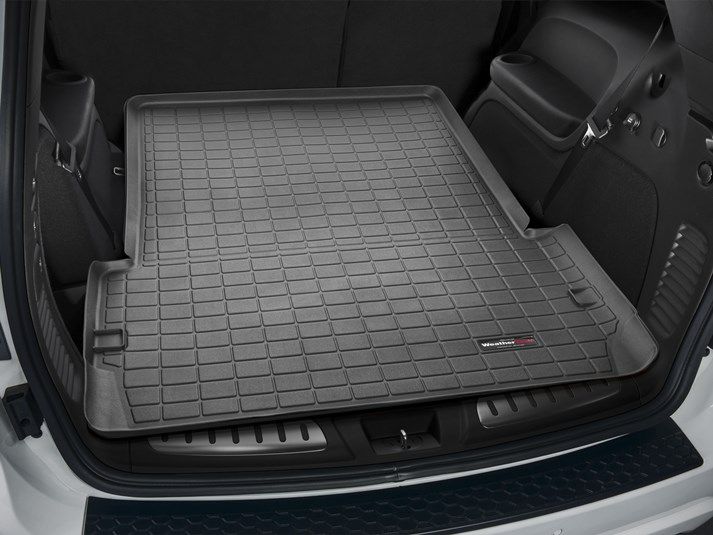}
    \end{subfigure}		
        \begin{subfigure}{0.23\linewidth}
      \centering
      \includegraphics[width=1\columnwidth]{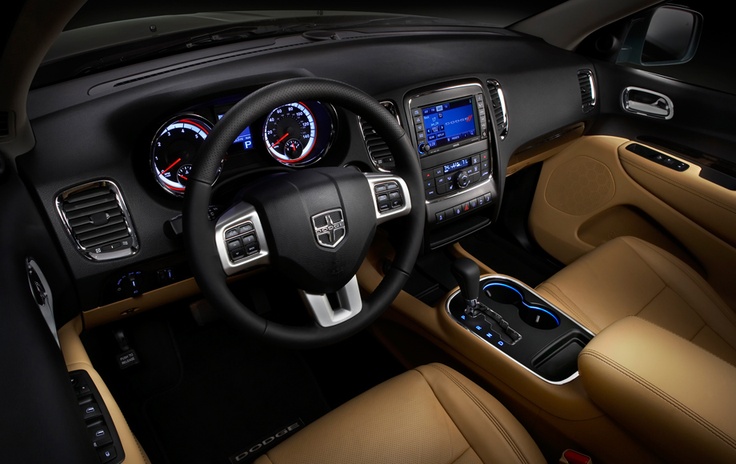}
    \end{subfigure}	
        \begin{subfigure}{0.23\linewidth}
      \centering
      \includegraphics[width=1\columnwidth]{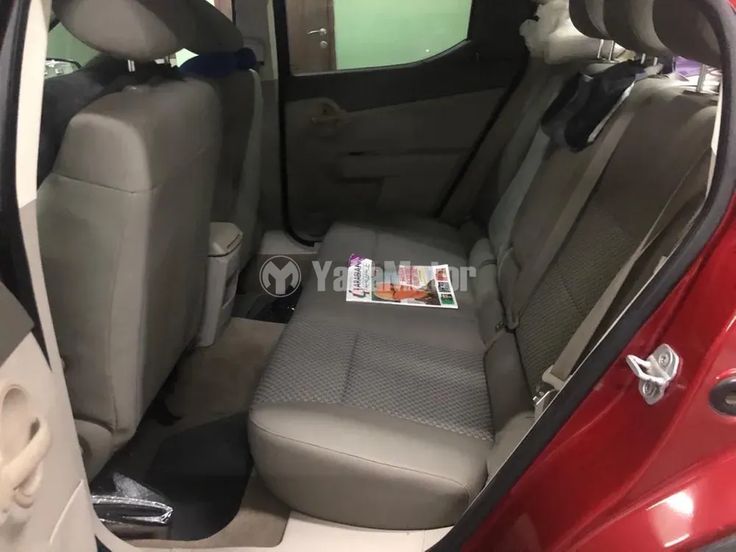}
    \end{subfigure}	
	\caption{Car interior images found in the class Dodge Durango SUV 2012 in CARS-98N.}\label{fig:interior}
\end{figure*}
\begin{figure*}[h!]
\centering
    \begin{subfigure}{0.45\columnwidth}
      \centering
      \includegraphics[width=1\columnwidth]{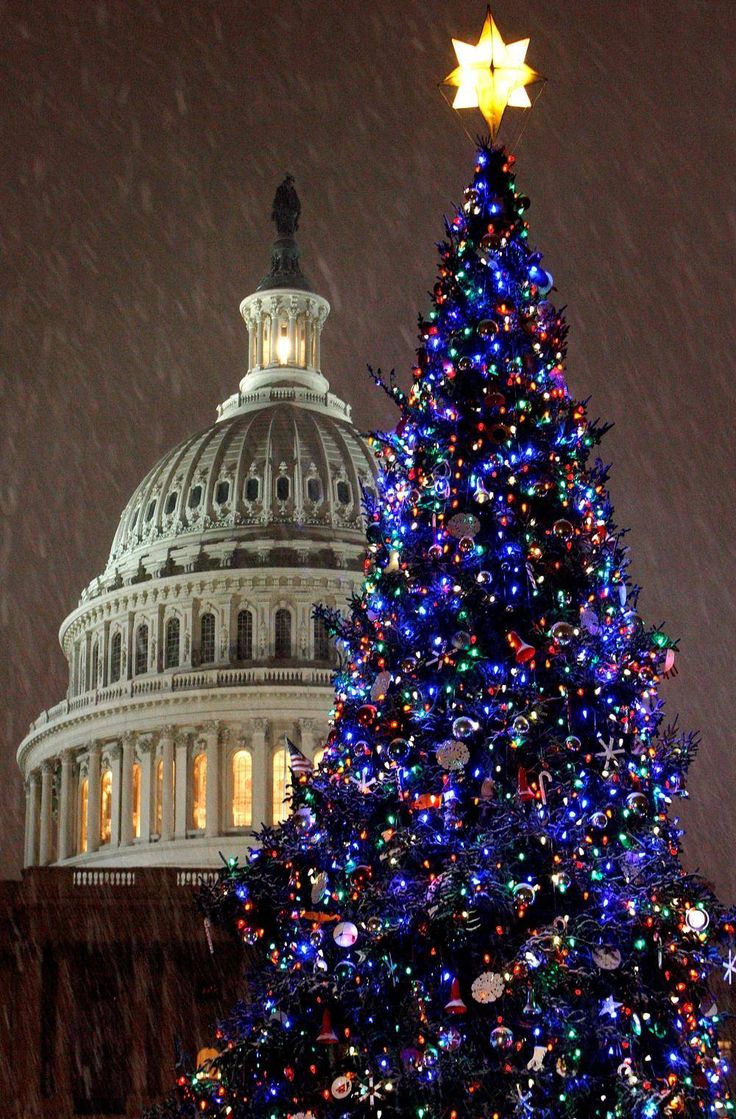}
    \end{subfigure}
    \begin{subfigure}{0.45\linewidth}
      \centering
      \includegraphics[width=1\columnwidth]{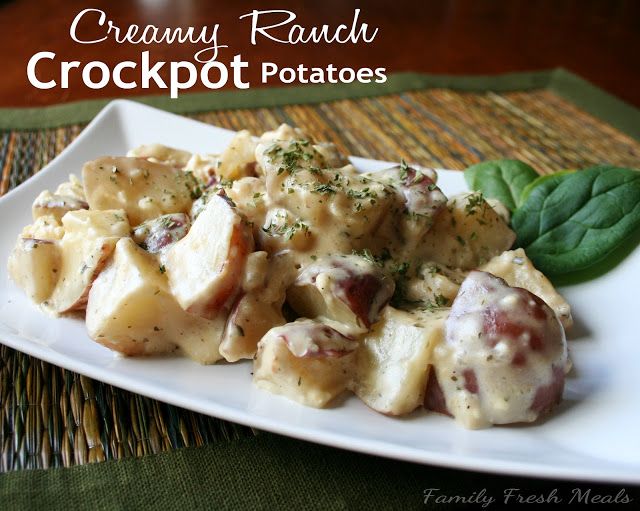}
    \end{subfigure}		
	\caption{Irrelevant images found in the class Eagle Talon Hatchback 1998 in CARS-98N.}\label{fig:other}
\end{figure*}

Figures~\ref{fig:othercar} to \ref{fig:interior} illustrate common types of noise in CARS-98N dataset. We take the class Dodge Durango SUV 2012 as an example. The noisy data contain images of different cars, as well as images of car parts and interior. As reference, we show the correct images in Figure~\ref{fig:correct}. We also observe that for small classes, the noisy data are often unrelated to cars. For example, Figure~\ref{fig:other} shows the noisy data in the class Eagle Talon Hatchback 1998.

{\small
\bibliographystyle{ieee_fullname}
\bibliography{supplement}
}